\documentclass[10pt,twocolumn,letterpaper]{article}

\usepackage{cvpr}
\usepackage{times}
\usepackage{epsfig}
\usepackage{graphicx}
\usepackage{amsmath}
\usepackage{amssymb}
\usepackage{subfig}
\usepackage{authblk}

\usepackage[breaklinks=true,bookmarks=false]{hyperref}

\cvprfinalcopy % *** Uncomment this line for the final submission

\def\cvprPaperID{732} % *** Enter the CVPR Paper ID here

\begin{document}
	
	%%%%%%%%% TITLE
	\title{Visual Question Reasoning on General Dependency Tree}
	
	\author{Qingxing Cao \quad\quad Xiaodan Liang  \quad\quad Bailing Li \quad\quad Guanbin Li \quad\quad Liang Lin\thanks{Corresponding author is Liang Lin. This work was supported by the State Key Development Program under
			Grant 2016YFB1001004, the National Natural Science Foundation of China under Grant 61622214 and Grant 61702565, Guangdong
			Natural Science Foundation Project for Research Teams under Grant 2017A030312006, and was also sponsored by CCF-Tencent Open Research Fund.}\\\vspace{2mm}
		School of Data and Computer Science, Sun Yat-sen University, China\\
		{\tt\small caoqx@mail2.sysu.edu.cn}, {\tt\footnotesize xdliang328@gmail.com}, {\tt\footnotesize liblin3@mail2.sysu.edu.cn}, {\tt\small liguanbin@mail.sysu.edu.cn, linliang@ieee.org}
		\vspace{-3ex}
	}
	
	\maketitle
	\thispagestyle{empty}

	%%%%%%%%% ABSTRACT
	\begin{abstract}
		The collaborative reasoning for understanding each image-question pair is very critical but under-explored for an interpretable Visual Question Answering (VQA) system. Although very recent works also tried the explicit compositional processes to assemble multiple sub-tasks embedded in the questions, their models heavily rely on the annotations or hand-crafted rules to obtain valid reasoning layout, leading to either heavy labor or poor performance on composition reasoning. In this paper, to enable global context reasoning for better aligning image and language domains in diverse and unrestricted cases, we propose a novel reasoning network called Adversarial Composition Modular Network (ACMN). This network comprises of two collaborative modules: i) an adversarial attention module to exploit the local visual evidence for each word parsed from the question; ii) a residual composition module to compose the previously mined evidence. Given a dependency parse tree for each question, the adversarial attention module progressively discovers salient regions of one word by densely combining regions of child word nodes in an adversarial manner. Then residual composition module merges the hidden representations of an arbitrary number of children through sum pooling and residual connection. Our ACMN is thus capable of building an interpretable VQA system that gradually dives the image cues following a question-driven reasoning route and makes global reasoning by incorporating the learned knowledge of all attention modules in a principled manner. Experiments on relational datasets demonstrate the superiority of our ACMN and visualization results show the explainable capability of our reasoning system.
		
	\end{abstract}
	
	%%%%%%%%% BODY TEXT
	\section{Introduction}
	
	\begin{figure}[t]
		\includegraphics[width=0.45\textwidth]{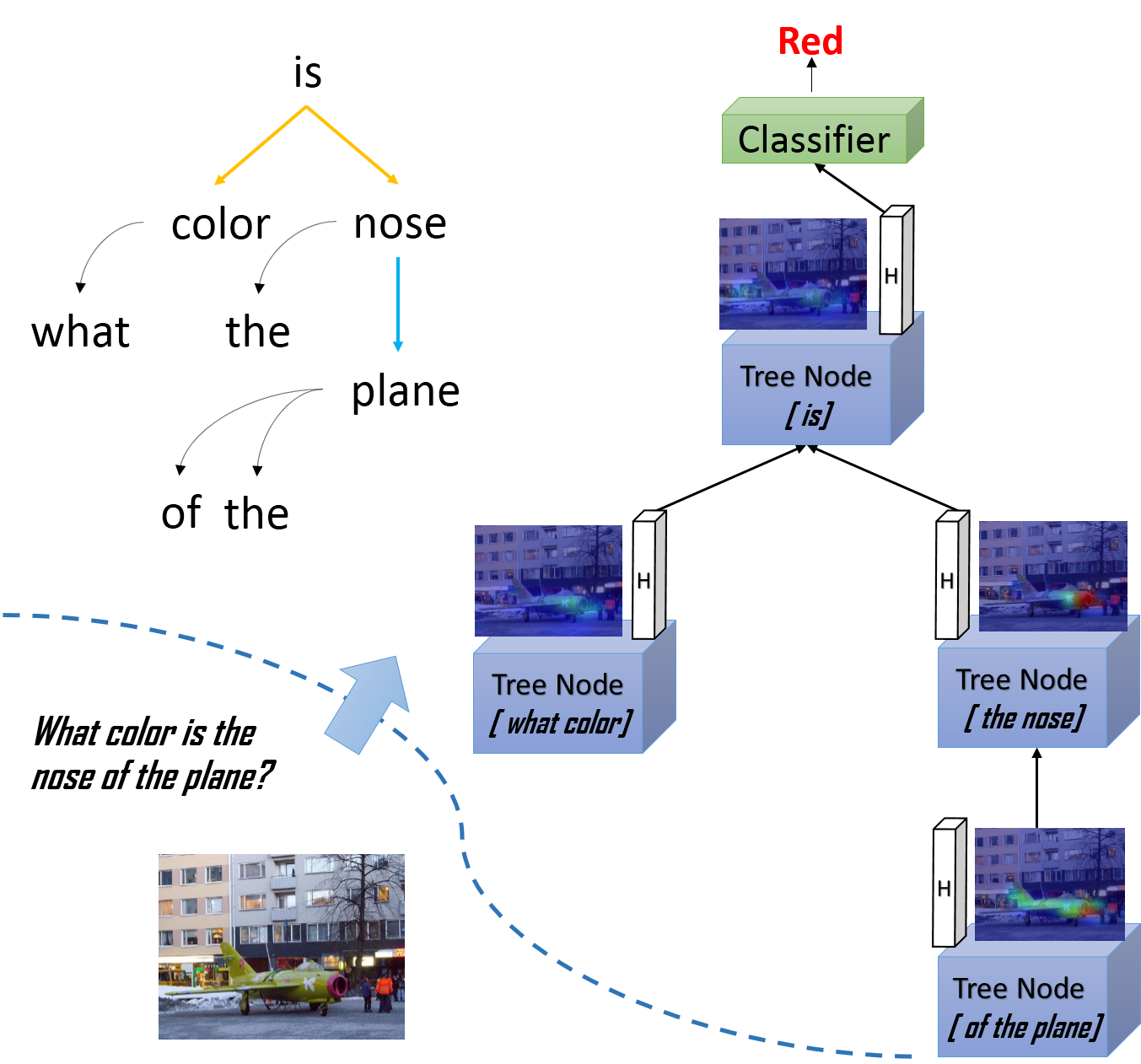}
		\caption{Illustration of our Adversarial Composition Module Network (ACMN) that sequentially performs reasoning over a dependency tree parsed from the question. Conditioning on preceding word nodes, our ACMN alternatively mines visual evidence for nodes with modifier relations via an adversarial attention module and integrates features of child nodes of nodes with clausal predicate relation via a residual composition module.
		}
		\vspace{-5mm}
		\label{fig:intro}
	\end{figure}
	
	The task of Visual Question Answering (VQA) is to predict the correct answer given an image and a textual question. 
	The key to this task is the capability of co-reasoning over both image and language domains.
	However, most of the previous methods~\cite{Seq2Seq, HiCoAtt, MLB} work more like a black-box manner, i.e., simply mapping the visual content to the textual words by crafting neural networks. The main drawback of these methods is the lack of interpreting ability to the results, i.e., why these answers are produced?
	Moreover, it has been shown that their accuracy may be achieved by over-fitting the data bias in the VQA benchmark~\cite{balanced_vqa_v2}, and the absence of explicitly exploiting structures of text and image leads to unsatisfying performance on relational reasoning~\cite{clevr}.
	Very recently, a few pioneering works~\cite{e2emn,inferring2017, GraphVQA} take advantage of the structure inherently contained in text and image, which parses the question-image input into a tree or graph layout and assembles local features of nodes to predict the answer. For example, layout ``\emph{more}(\emph{find}(\emph{ball}), \emph{find}(\emph{yellow}))" means the module should locate the ball and the yellow object on the image first, then compose the two results to answer whether there are more balls than yellow objects.
	However, these methods would either rely on hand-designed rules for understanding questions or train a layout parser from scratch which suffers large decay in performance. We argue those limitations severely prohibit their application potentials in understanding general image-question pairs that may contain diverse and open-ended question styles.

	To achieve a general and powerful reasoning system with the ability to enable reasoning over any dependency trees of questions rather than fixed layouts in prior works, we propose a novel  Adversarial Composition Modular Network (ACMN) that designs two collaborative modules to perform tailored reasoning operations for addressing two most common word relations in the questions. As shown in Figure~\ref{fig:intro}, given a specific dependency tree of each question by an off-the-shelf dependency parser, we construct a reasoning route following the dependency layout that is a tree-structure composed of clausal predicate relation and modifier relation. Our module network then alternatively performs two collaborative modules on each word node for global reasoning: 1) exploit local visual evidence of each word guided by exploiting regions of its child nodes in an adversarial way in terms of nodes with modifier relations; 2) integrate the hidden representations of child nodes via residual composition with respect to nodes with clausal predicate relation. Notably,
	in contrast to previous methods, our ACMN aims at a general and interpretable reasoning VQA framework that does not require any complicated handcrafted rules or ground-truth annotation to obtain a specific layout.
	
	Specifically, we observe that the frequently used types of dependency relations can be categorize in two sets: whether the head is a predicate that describes the relation of its children (\eg \emph{color} $\leftarrow$ \emph{is}, \emph{is}$\rightarrow $\emph{nose}), or a word decorated by its child (\eg \emph{furthest}$\rightarrow$\emph{object}). We refer the first set as clausal predicate relation and the second is modifier relation. Our ACMN designs adversarial attention modules for encoding modifier relation and residual composition modules for clausal predicate relations. 
	
	Firstly, for child nodes with modifier relations, we apply the adversarial attention mechanism similar to \cite{AdvErase}. To enable effectively mining all visual evidence, we enforce each parent node explore new regions by masking out attentive regions of its child nodes at each step. More specifically, we sum up the attention maps from child nodes and mask out features weighted by the mined attention map in a soft manner. We then perform attention operation on manipulated hidden representations to extract new local visual evidence for the parent node.
	Secondly, for those child nodes with clausal predicate relation, our residual composition module integrates the hidden representations weighted by attention maps of its child nodes using bilinear fusion. In order to retain the information from the child nodes and deal with an arbitrary number of child nodes, the module learns a residual that will be added to the input on the sum of child nodes to modify their hidden representations.
	Finally, the final hidden representation of the root node goes through a multi-layer perceptron to predict the final answer. 
	
	Extensive experiments show that our model can achieve state-of-art VQA performance on both natural image VQA benchmark VQAv2 dataset and CLEVR relational dataset. And qualitative results further demonstrate the interpretable capability of our ACMN on collaborative reasoning over image and language domains.
	
	Our contributions are summarized as follows: 1) We present a general and interpretable reasoning VQA system following a general dependency layout composed by modifier relations and clausal predicate relations. 2) a novel adversarial attention module is proposed to enforce efficient visual evidence mining for modifier relations while a residual composition module for integrating knowledge of child nodes for clausal predicate relations.
	
	\begin{figure*}[t!]
		\centering
		%  \subfloat[h][]{
		%   \label{fig:improved_subfig_a}
		%  % \begin{minipage}[t]{0.3\textwidth}
		%    \centering
		%    \includegraphics[height=0.08\textheight]{figure/intro.png}
		%  % \end{minipage}
		%  }
		\begin{minipage}[b]{0.53\textwidth}
			\subfloat[h][Adversarial Composition Neural Module]{
				\label{fig:block}
				\centering
				\includegraphics[width=1.0\textwidth]{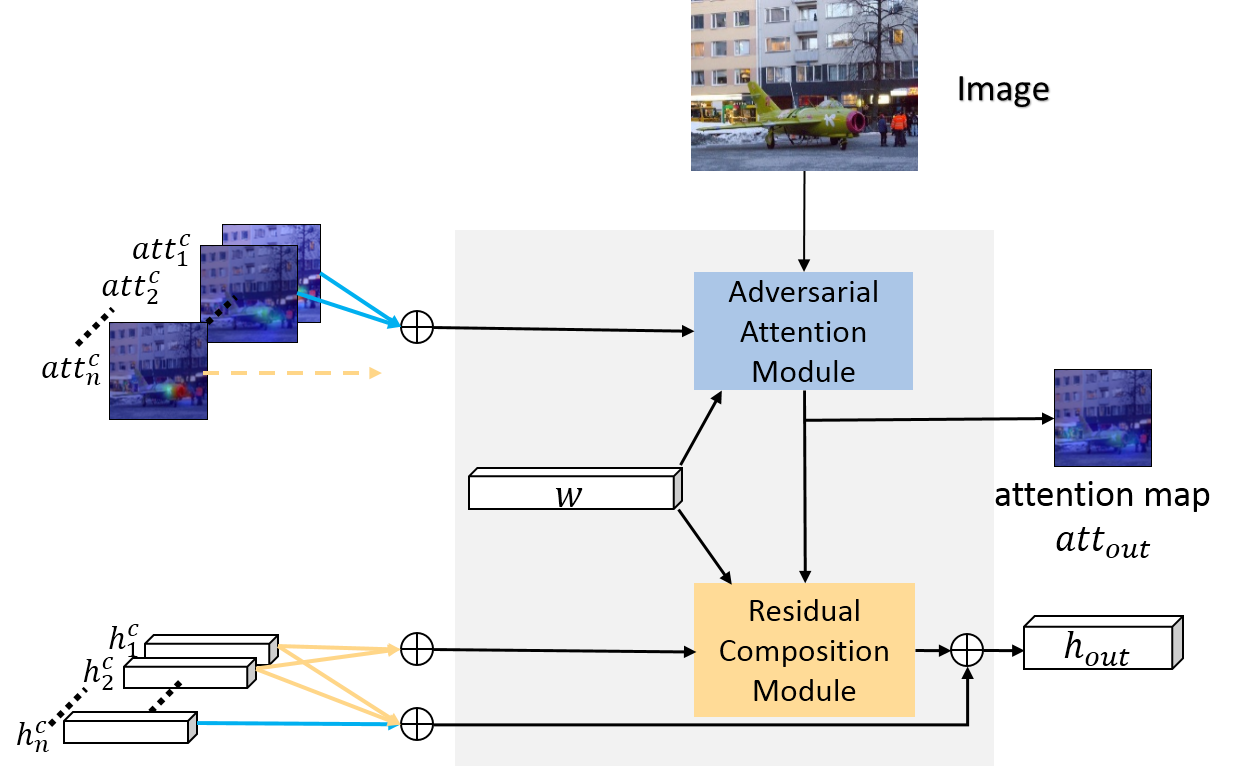}
			}
		\end{minipage}
		%\hspace{0.1mm}
		\hfill
		\begin{minipage}[b]{0.46\textwidth}%[s]{0.40\textwidth}
			\subfloat[][Adversarial Attention Module]{
				\label{fig:att}
				\includegraphics[width=1.0\textwidth]{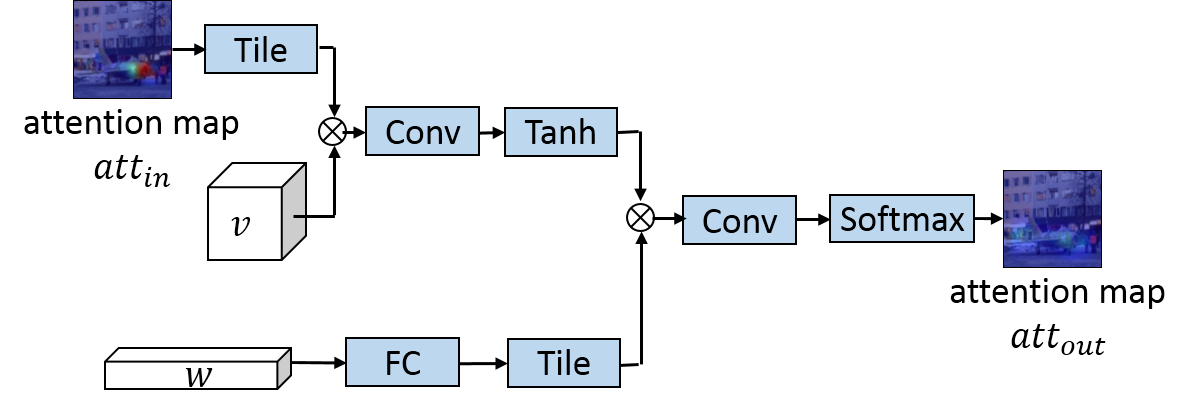}
			}
			\vfill
			\subfloat[][Residual Composition Module]{
				\label{fig:h}
				\includegraphics[width=1.00\textwidth]{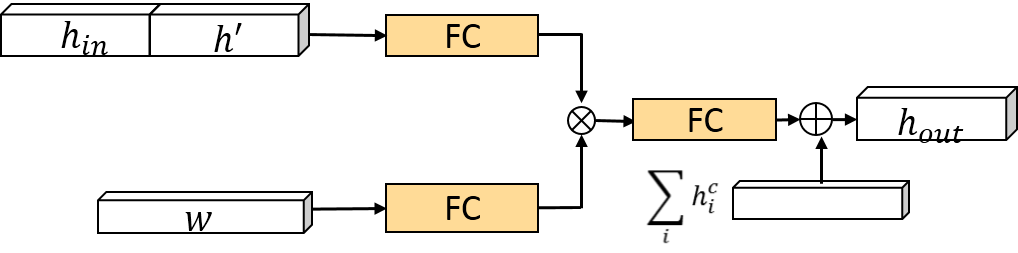}
			}
		\end{minipage}
		\caption{The modules in our ACMN: a) each ACMN module is composed of an adversarial attention module and a residual composition module; b) adversarial attention module; c) residual composition module. The blue arrows indicate the modifier relation and the yellow arrows represent the clausal predicate relation. Each node receives the output attention maps and the hidden features from its children, as well as the image feature and word encoding. The adversarial attention module is employed to generate a new attention map conditioned on image feature, word encoding and previously attended regions given by modifier-dependent children. The residual composition module is learned to evolve higher-level representation by integrating features of its children and local visual evidence.}\vspace{-5mm}
	\end{figure*}
	
	\section{Related Works}
	\paragraph{Visual question answering} The visual question answering task requires co-reasoning over both image and text to infer the correct answer.
	
	The baseline method proposed in VQA dataset~\cite{VQA} to solve this task uses a CNN-LSTM based architecture, which consist of a convolution neural network to extract image features, and a LSTM for literal question feature encoding. It combines these two features to predict the final answer. In recent years, a large number of works followed this pipeline and have achieved substantial improvements over the baseline model. Among these works, the attention mechanism~\cite{DBLP:journals/corr/IlievskiYF16,Xu2016,shih2016att,Visual7W,StackedAtt,HiCoAtt} and the joint embedding of image and question representation~\cite{MCB,MLB} have been widely studied. Attention mechanism learns to focus on the most discriminate sub-region instead of the whole image, provides a certain extent of reasoning to the answer. Different attention methods such as stacked attention~\cite{StackedAtt} and co-attention between question and image on different levels~\cite{HiCoAtt} constantly improve the performance of the VQA task. As for the multi-modal joint embedding, Fukui \emph{et~al.}~\cite{MCB}, Kim \emph{et~al.}~\cite{MLB} and Hedi \emph{et~al.}~\cite{MUTAN} exploited the compact bilinear method to fuse the embedding of image and question and incorporated the attention mechanism to further improve the performance.
	
	However, some recently proposed works~\cite{balanced_vqa_v2,ECCV16baseline} showed that the promising performance of these deep models might be achieved by exploiting the dataset bias. It is possible to perform equally well by memorizing the QA-pairs or encode the question with the bag-of-words method. To address this concern, newer datasets were released in the very recent. The VQAv2 dataset~\cite{balanced_vqa_v2} was proposed to eliminate the data biases through balancing question-answer pairs. The CLEVR~\cite{clevr} dataset consists of synthetic images, and provides more complex questions that involve multiple objects. It also has balanced answer distribution to suppresses the data bias.
	
	\textbf{Reasoning model}
	There exist prior works that tried to explicitly incorporate the knowledge into the network structure. \cite{externVQA, FVQA} encoded both image and question into discrete vectors such as image attributes or database queries. These vectors enable their model to query the external data source for common senses and basic factual knowledge to answer the question. \cite{zhu2017cvpr} actively acquires pre-defined types of evidence to obtain external information and predicted the answer. Other recent works proposed modular network to handle the composition reasoning. \cite{DynaMem} augmented a differentiable memory and encoded long-term knowledge for answer inference.
	
	%\cite{structureAtten}
	
	\textbf{Neural modular network}
	The recently proposed neural modular network provides a framework to address compositional visual reasoning. 
	Instead of using a fixed structure to predict the answer to every question, this line of works assembles a structure layout for different question into pre-defined sub-tasks. 	Then a set of neural modules is designed to solve a particular sub-task respectively. Earlier works~\cite{nmn,lnmn} generated their layouts based on dependency parser. Later, \cite{e2emn,inferring2017} use sequence-to-sequence RNN to predict the post-order of layout tree, and jointly train the RNN and the neural module using RL or EM scheme.
	
	However, it is difficult to jointly train the RNN and the modular network from scratch. On the other hand, existing neural module may propagate its error through the rest of modular network, thus these methods heavily depend on the correctness of the structured layout. Our method instead modifies the structure of neural modules to avoid the error propagation and takes advantage of the information laying on the type of dependency. This substantially improves the prediction accuracy while preserving the compositional reasoning ability.

	\section{Adversarial Composition Modular Network}

	\subsection{Overview}
	Given the free-form questions $Q$ and images $I$, our proposed ACMN model learns to predict the answers $y$ and their corresponding explainable attention maps.
	%$z$: $F(z, y | I, Q)$. 
	Specifically, we first generate the structure layout given the input question $Q$ by parsing it into a tree structure using an off-the-shelf universal Stanford Parser~\cite{dep}. To reduce the computational complexity, we prune the leaf-nodes that are not noun, then categorize the labels of dependency relations such as ``nominal modifier" (\eg (\emph{left}, \emph{object})), ``nominal subject"(\eg (\emph{is}, \emph{color}) into two classes: the modifier relation $M$ and the clausal predicate relation $P$.% The detailed of two categories is shown in Table~\ref{table:UD}.
	%The resulting layout is shown as Figure~\ref{fig:core}.
	
	Our ACMN model is constituted by a set of network modules $f$ on each word node in the layout from bottom to top.
	Suppose a node is $x$, and its $n$ children are denoted as $\{x^c_1,x^c_2,...,x^c_n\}$. 
	%we have $[att_{out},h_{out}] = f(v,w,X^c).$
	The module $f$ has three inputs: the image feature $v$, the word encoding $w$, and its children's outputs $[att^{c}_i,h^{c}_i] = f(x^c_i)$. It outputs a new attentive region $att_{out}$, and a hidden feature $h_{out}$, which are generated by the adversarial attention module $f_a$ and residual composition module $f_h$ respectively, as shown in Figure~\ref{fig:block}.
	
	The spatial feature $v$ is extracted for each image via any pre-trained convolution neural network on ImageNet (\eg conv5 features from ResNet-152~\cite{DBLP:conf/cvpr/HeZRS16} or conv4 features from ResNet-101~\cite{DBLP:conf/cvpr/HeZRS16}). The word embedding vector $w$ is obtained with a Bi-LSTM~\cite{BiLSTM}. Specifically, each word in the question is first embedded as a $300$ dimension vector, then the question is fed into a bidirectional LSTM. The final word embedding $w$ is the hidden vector of Bi-LSTM at its corresponding position.
	
	\subsection{Adversarial Attention Module} Specifically, as shown in Figure~\ref{fig:att}, we first filter the child nodes whose relation is modifier $M$ and perform adversarial attention module on the parent node $x$. The input attention map $att_{in}$ of each node $x$ is first obtained by summing attention maps $\{{att}^c_i\}$ of its modifier-dependent child nodes $x_i$. The adversarial mask is generated by subtracting $att_{in}$ by $1$ followed by a ReLU layer to keep the results non-negative. Then the mask is used to softly weight the spatial feature $v$ via a multiplication operation. Finally, the adversarial module $f_a$ outputs a new attention map ${att}_{out}$ conditioned on the input word embedding $w$ and weighted spatial features. We further apply Softmax to regularize the resulting attention map into the range of $[0,1]$. The visual representation $h'$ of the node $x$ is then generated by the weighted sum of each grid features in $v$ given the attended weight ${att}_{out}$.

	\subsection{Residual Composition Module}
	As shown in Figure~\ref{fig:h}, the residual composition module $f_h$ first sums the hidden features $\{{h}^c_i\}$ of its children with clausal predicate relation $P$ into $h_{in}$, and then concatenate $h_{in}$ with extracted local evidence $h'$, and finally combine with word encoding $w$ to generate a new hidden feature $h_{out}$. A fully connected layer is applied to project both the concatenated hidden $[h_{in}, h']$ and word encoding $w$ feature to $2048$ dimension feature vector. Then we perform element-wise multiplication on the two features, project it to $128$ dimension vector, and add it with all of its children's hidden feature $\{{h}^c_i\}$ as the output hidden representation $h_{out}$.

	\begin{table}[t] \centering \small%\footnotesize %
		\center
		\begin{tabular}{|*{2}{c|}}
			\hline
			Clausal Predicate Relation & Relation Description \\
			\hline\hline
			NSUBJ & Nominal subject \\
			NSUBJPASS & Passive nominal subject \\
			CSUBJ &  Clausal subject\\
			CSUBJPASS & Clausal passive subject\\
			DOBJ & Direct object \\
			IOBJ & Indirect object \\
			CCOMP & Clausal complement \\
			XCOMP & Open clausal complement \\
			\hline
			\hline
			Modifier Relation & Relation Description \\
			\hline\hline
			NMOD & Nominal modifier \\
			AMOD & Adjectival modifier \\
			NUMMOD & Numeric modifier \\
			ADVMOD & Adverbial modifier\\
			APPOS & Appositional modifier \\
			ACL & Clausal modifier of noun\\
			DET & Determiner \\
			CASE & Prepositions, postpositions... \\ %and other case markers 
			COMPOUND & Compound\\
			\hline
		\end{tabular}
		\caption{The two major categories of relations classified by the universal dependency parser.}
		\label{table:UD}
		\vspace{-4mm}
	\end{table}
	
	\subsection{The proposed ACMN model}
	
	Given the tree-structured layout of the dependency tree, our ACMN module is sequentially used on each word node to mine visual evidence and integrate features of its child nodes from bottom to top, and then predict the final answer at the root of the tree.
	Formally, each ACMN module can be written as:
	\begin{equation}
	\begin{aligned}
	{att}_{in} &= \sum_{(x,x^c_i) \in M}{att}^c_{i},  \\
	h_{in} &= \sum_{(x,x^c_i) \in P}h^c_{i}, \\
	{att}_{out}& = f_{a}({att}_{in}, v, w), \\
	h' &= {att}_{out} * v, \\
	h_{out} &= f_h([{h}_{in}, h'], w) + \sum_{i} h^c_{i}, \\
	\end{aligned}
	\label{eq:global}
	\end{equation}
	Where $(x,x^c_i)$ represents the relation of node $x$ and its child $x^c_i$.
	Because the nodes with modifier relations $M$ can modify their parent node by referring to a more specific object, we thus generate a more precise attention map as ${att}_{out}$. On the other hand, the clausal predicate relation $P$ suggests the parent node is a predicate of child nodes, we thus integrate features of child nodes to enhance the representation given the predicate word.
	
	After propagating through all word nodes with a sequence of adversarial attention module and residual composition module, the output features of the root node $h_{root}$ are passed through a three Multi-Layer Perceptron to predict the final answer $y$. Our model that is stacked by a list of adversarial attention modules and residual composition modules following a tree-structured layout. Weights are share across modules with same height in order to learn different levels of semantic representation. The whole model can be trained end-to-end with only the supervision signal $y$.

	\subsection{Modifier Relation and Clausal Predicate Relation}
\begin{figure}[t]
		\centering
		\includegraphics[width=0.45\textwidth]{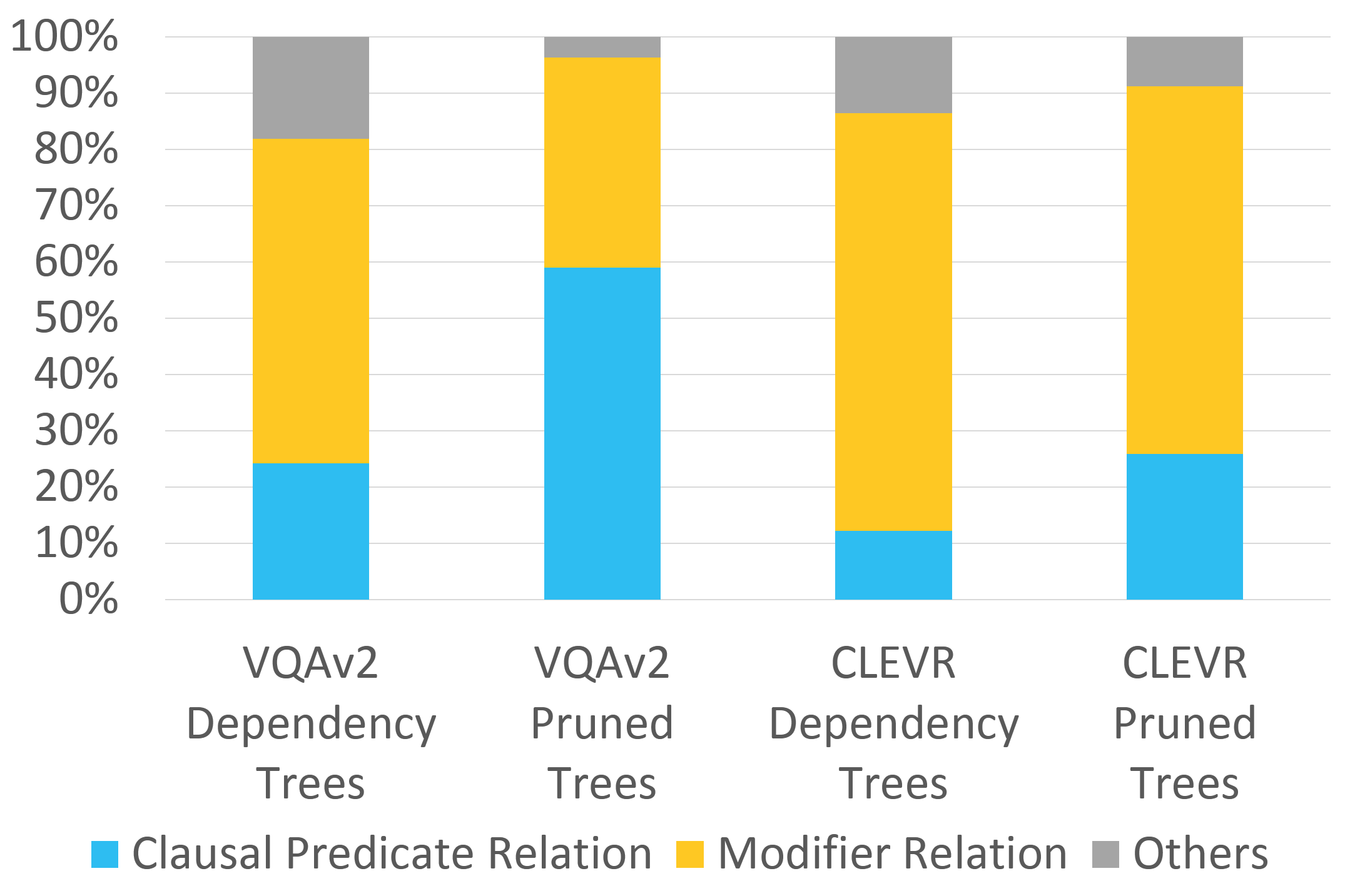}
		
		\caption{The statistic of clausal predicate relation and modifier relation in the questions of VQAv2~\cite{balanced_vqa_v2} dataset training split and CLEVR dataset~\cite{clevr} training split.
		}
		\label{fig:static}
		\vspace{-5mm}
	\end{figure}

	A dependency-based parser is to draw directed edges from head words to dependent words in a sentence. It also labels the head-dependent relations to provide an approximation to the relationship between predicates and their arguments. One of the most widely used head-dependent relation sets is the Universal Dependencies(UD)~\cite{UD}. It has a total of 42 relations that can be clustered into 9 categories. But the frequently used relations concentrate on only two of them: the core dependents of clausal predicate and the noun dependents, as shown in Figure~\ref{fig:static}. In this work, we make some small modification to noun dependents sets and refer these two kinds of relationships as clausal predicate relation $P$ and modifier relation $M$. The details of both sets are shown on Table~\ref{table:UD}. For those relations that belong to neither of the two sets, we will pass both the attention map and the hidden representation to the parent nodes.
	
	Dependents of clausal predicate relation $P$ describe syntactic roles with respect to a predicate that often describes how to compose its children.
	For example, in question \emph{What color is the nose of the plane?}, word \emph{is} is the head of \emph{color} and \emph{nose}, and their relations are \emph{(is,color) = direct object} and \emph{(is,nose) = nominal subject}. So the word \emph{is} tells us how to compose word \emph{color} and \emph{nose}, such as using a function ``describes(\emph{color, nose})" in a modular network~\cite{nmn}. Thus, our residual composition module learns to compose features $\{h^c_i\}$ of its children nodes with clausal predicate relation $P$ conditioned on current word embedding $w$ of a parent node.
	
	The modifier relations $M$ categorize the ways words that can modify their parents. For example, the modifier relation $M$ of the question \emph{What size is the cylinder that is left of the brown metal thing} from CLEVR dataset~\cite{clevr} can be the relation \emph{(left, brown metal thing) = nominal modifier}. The reason is that the word \emph{left} indicates the region related to \emph{brown metal thing} instead of \emph{cylinder}, which is similar to ``transform(\emph{left, thing})" relation in the modular network~\cite{nmn}. Thus we can obtain a  modified attention map for the part node according to attention maps $\{{att}^c_i\}$ of its children given the current word encoding $w$ via our adversarial attention module.

	\section{Experiment}

	\begin{figure*}[t]
		\centering
		\includegraphics[width=1.0\textwidth]{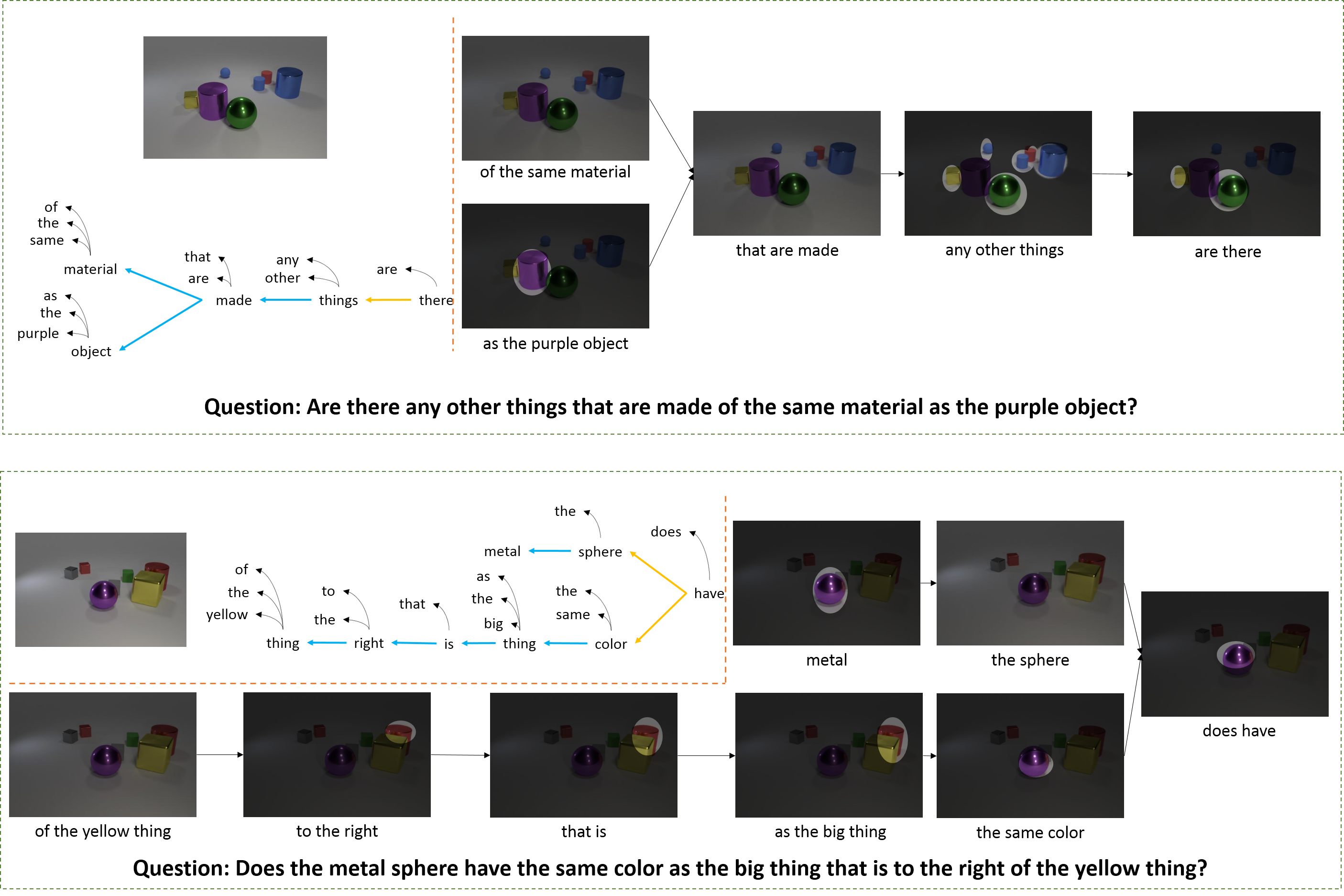}
		
		\caption{Two examples of the dependency trees of questions and corresponding regions attended by our model at each step on CLEVR dataset. The question is shown on the bottom. The image and dependency parse tree are shown on the left. The arrows in the dependency tree are drawn from the head words to the dependent words, The blue arrows indicate the modifier relation $M$, and the yellow arrows indicate the clausal Predicate relation $P$. The curved arrows point to the pruned leaf words that are not a noun. Thus word ``there" and ``have" is the root node for each example respectively. The regions with high attention weight are shown as bright areas in the images on the right. Those nodes without obvious bright region indicate our model equally attend all regions of the image, thus no specific salient regions correspond to this node. }\textbf{}
		\label{fig:clevr}
		\vspace{-4mm}
	\end{figure*}
	
	We validate the effectiveness and interpretation capability of our models on both two synthetic datasets (i.e., CLEVR and Sort-of-CLEVR) that mainly focus on relation reasoning and one natural dataset (i.e., VQAv2) with diverse image-question pairs in the wild.
	
	\subsection{Datasets}
	
	The \textbf{CLEVR}~\cite{clevr} is a synthesized dataset with $100,000$ images and $853,554$ questions. The images are photo-realistic rendered images with objects of random shapes, colors, materials and sizes. The questions are generated using sets of functional programs, which consists of functions that can filter certain color, shape, or compare two objects. Thus, the reasoning routes required to answer each question can be precisely determined by its underlying function program. Unlike natural image dataset, it requires model capable of reasoning on relations to answer the questions. %this dataset explicitly demands to reason on relations by asking a large number of questions on object pairs.

	The \textbf{Sort-of-CLEVR}~\cite{RelNet} consists of synthesized images of 2D colored shapes.
	Each image has exactly 6 objects that can be unambiguously identified by 6 colors, and the objects have random shapes(square or circle) and positions. Each image is associated with 20 questions asking about the shape or position of a certain object, 10 of which is non-relational questions that query the object by its unambiguous colors and another 10 are relational questions that query the object with furthest or closest relation to another unambiguous colored object.
	It is visually simpler than the CLEVR, but also requires the model capable of relational reasoning. Since the original dataset is not released, we generate a set following their detailed description, including $9800$ images for training and $200$ for testing. 
	
	\begin{table*}[th!]
		\centering %\small%\footnotesize %
		\resizebox{\textwidth}{!}{%
			\begin{tabular}{|c|cc|ccc|cccc|cccc|c|}
				\hline
				& & &\multicolumn{3}{c|}{Compare Integer} & \multicolumn{4}{c|}{Query} & \multicolumn{4}{c|}{Compare} & \\
				\hline
				Method  &Exist &Count & Equal& Less& More & Size& Color&Material&Shape&Size&Color& Material& Shape & Overall\\
				\hline
				
				LBP-SIG~\cite{SAVQA} & 79.63& 61.27 & \multicolumn{3}{c|}{80.69} & \multicolumn{4}{c|}{88.59} & \multicolumn{4}{c|}{76.28} & 78.04 \\
				RN~\cite{RelNet} & 97.8 & 90.1 & \multicolumn{3}{c|}{93.6} & \multicolumn{4}{c|}{97.9} & \multicolumn{4}{c|}{97.1}  & 95.5\\
				N2NMN scratch~\cite{e2emn}& 72.7& 55.1& 71.6& 85.1 &79.0& 88.1 &74.0 &86.6 &84.1 &50.1& 53.9 &48.6& 51.1 & 69.0  \\
				N2NMN cloning expert~\cite{e2emn}  & 83.3 & 63.3 & 68.2 & 87.2& 85.4 & 90.5& 80.2& 88.9& 88.3 & 89.4& 52.5& 85.4& 86.7 & 78.9 \\
				N2NMN policy search~\cite{e2emn}& 85.7& 68.5 & 73.8& 89.7& 87.7 & 93.1& 84.8& 91.5& 90.6 & 92.6& 82.8& 89.6& 90.0  & 83.7\\
				PE-semi-9K~\cite{inferring2017} & 89.7& 79.7 &85.2& 76.1& 77.9 & 94.8& 93.3& 93.1& 89.3 & 97.8& 94.5& 96.6& 95.1& 88.6 \\
				PE-Strong~\cite{inferring2017}& 97.7& 92.7 & 98.0& 99.0& 98.9 & 98.8& 98.4& 98.1& 97.3 & 99.8& 98.5& 98.9& 98.4  & 96.9\\
				
				\textbf{Ours}  & 94.21 & 81.37 & 75.06 & 88.23 & 81.51 & 92.61 & 86.45 & 92.35 & 90.65 & 98.50 & 97.44 & 94.93 & 97.37& 89.31 \\
				\hline
			\end{tabular}
		}
		\caption{Comparisons in terms of question answering accuracy on the CLEVR dataset. The performance of question types \emph{Exist},\ emph{Count}, \emph{Compare Integer}, \emph{Query}, \emph{Compare} are reported on each column. LBP-SIG~\cite{SAVQA} and RN~\cite{RelNet} only report total accuracy of question types \emph{Compare Integer}, \emph{Query}, \emph{Compare}, their performance on these types are mereged.
		}\vspace{-4mm}
		\label{table:clevr}
	\end{table*}	
	
	The \textbf{VQAv2}~\cite{balanced_vqa_v2} contains $204,721$ natural images from COCO~\cite{Lin2014} and $1,105,904$ free-form questions. Compared with its first version~\cite{VQA}, this dataset focuses on reducing dataset biases through balanced pairs: for each question, there are pair of images which the answers to that question are different.
	
	\subsection{Implementation details}
	For the CLEVR dataset, we employ the same setting used in \cite{SAVQA,clevr} to extract the image feature and words encoding. We first resize all images to $224 \times 224$, then extract the conv4 feature from ResNet-101 pre-trained on ImageNet. The resulting $1024 \times 14 \times 14$ feature maps are concatenated with a 2-channel coordinate map. It is further fed into a single $3 \times 3$ convolution layer. The resulting $128 \times 14 \times 14$ feature maps are passed through our ACMN module network. We encode the questions using a bidirectional LSTM~\cite{BiLSTM} with $1024$-d hidden states for both directions. The hidden vector of Bi-LSTM at corresponding word position is considered as this word's encoding $w$. The maximum tree height of this dataset is $13$, thus there a total of $13$ node module instances. The hidden representation $h$ of each module is a $256$-d vector. The three-layer MLP have output sizes of $512$, $1024$ and $29$ respectively. 
	
	The size of images in Sort-of-CLEVR is $75 \times 75$, we used a four-layer CNN and each layer has a $3 \times 3$ kernel and $24$-channel outputs to extract the image features. The resulting feature maps have the size of $24 \times 8 \times 8$, so the output $h'$ of the adversarial attention module is a $24$-d vector. The output $h$ of the residual composition module is a $256$-d vector, and there are $5$ instances of modules for each level of dependency tree. The words in a question are first embedded as a $300$-d vector and then the whole question is encoded by a bidirectional LSTM~\cite{BiLSTM}, which has $150$-d hidden units in both directions. The word encoding vector is represented by the LSTM hidden vector at its corresponding position. 
	
	The image features for VQAv2 are extracted by bottom-up attention network~\cite{Anderson2017up-down}, which is trained to detect objects on the \textit{Visual Genome} dataset. The word vectors are also extracted from a $150$-d bidirectional LSTM at corresponding positions. The maximum height of dependency parse trees in this dataset is $11$. The hidden representation $h$ of these $11$ module instances are $1024$-d vectors.
	
	For VQAv2, We train our model on the training and validation split. For CLEVR and Sort-of-CLEVR dataset, only the training split is used. The model is trained with Adam optimizer~\cite{adam}. The base learning rate is $0.0001$ and the batch-size is $32$. The weight decay, ${\beta}_1$ and ${\beta}_2$ are $0$, $0.9$, $0.999$ respectively, which are the default settings for Adam optimizer.
	
	%\subsection{Evaluation metrics}
	%\xiaodan{TODO}
	
	\begin{table}[t] \centering \small%\footnotesize %
		\center
		\begin{tabular}{|*{3}{c|}}
			\hline
			& Non-relational & Relational \\
			\hline
			Ours-w/o residual & 99.05 & 93.50 \\
			Ours-DualPath~\cite{DPNet} & 98.05 & 91.10 \\
			\hline
			Ours-relocate~\cite{e2emn} & 98.20 & 90.10 \\
			Ours-concat & 99.10 & 91.15 \\
			%Fully connected shift& 99.10 & 93.90 \\
			\hline
			CNN+MLP~\cite{RelNet} & - & 63 \\ %99.0
			CNN+RN~\cite{RelNet} & - & 94.0 \\ %99.0
			\textbf{Ours} & \textbf{99.85} & \textbf{96.20} \\
			\hline
		\end{tabular}
		\caption{Comparisons in terms of question answering accuracy on the Sort-of-CLEVR dataset.}\vspace{-6mm}
		\label{table:sclevr}
	\end{table}
	
	\subsection{Comparison with State-of-the-arts}
	\subsubsection{CLEVR dataset}
	Table~\ref{table:clevr} shows the performances of different works on CLEVR test set. The previous End-to-End modular network~\cite{e2emn} and Program Execution Engine~\cite{inferring2017} are shorten as N2NMN and PE respectively. They both use the functional programs as groundtruth layout, and train their question parser with a sequence-to-sequence manner with strong supervision. They also have variants that are trained using semi or none supervision signals. The ``N2NMN scratch" indicating the end-to-end modular network without layout supervision and the ``N2NMN cloning expert" show the results of their model trained with full supervision. The ``N2NMN policy search" gives this model's best results if it further trains the parser from ``N2NMN cloning expert" with RL. It can be seen that our model outperforms all of these previous models by a large margin without using any dataset-specific layout, showing the good generalization capability of our ACMN. Our ACMN also beats the Program Execution Engine~\cite{inferring2017} variant trained with semi-supervision (as ``PE-semi-9K"). 
	The PE-Strong~\cite{inferring2017} used all program layouts as additional supervision signals, and the RN~\cite{RelNet} is a black-box model that lacks interpreting ability. Although our ACMN only obtains comparable results with Program Execution Engine~\cite{inferring2017} with fully-supervision (as ``PE-Strong") and Relation Network (as ``RN")~\cite{RelNet}, our ACMN can provide more explicit reasoning results without layout supervision.
	
	Figure~\ref{fig:clevr} shows the promising intermediate reasoning results achieved by our ACMN.
	The images and the dependency parse trees are shown on the left. We highlight the regions with high attention weights, and slightly brighter the image if our model equally attends all of the regions. The first example shows that our model can first locate the ``purple object", while the phase ``same material" alone does not correspond to any object, our model doesn't focus on any specific region. Later, our model attends all object except the purple object given phase ``any other things", and ``are there" locate the objects that have the same material and predict the answer ``yes". The second example illustrates the process of locating the ``big thing" and the ``metal sphere". Then our model composes their visual features to answer whether these two objects have the same color.
	
	\vspace{-4mm}
	\subsubsection{Sort-of-CLEVR dataset}
	
	Table~\ref{table:sclevr} shows the comparisons among our model, its variants and prior works on Sort-of-CLEVR dataset. As described in \cite{RelNet}, since the visual elements in this dataset are quite simple, a simple CNN+MLP baseline model can achieve over $94\%$ accuracy for non-relational questions but fail for relational questions. We thus mainly focus on comparing results for relational questions. The results of two baselines (i.e. ``CNN+MLP~\cite{RelNet}" and ``CNN+RN~\cite{RelNet}" are originally reported in~\cite{RelNet}. The actual accuracy number for non-relational questions are not reported since both models achieve nearly $100\%$. We can see that our ACMN achieves superior results over two previous methods for answering relational questions that require the model has strong capability in relation reasoning rather than overfitting the dataset bias as previous works.
	
\begin{figure}[t]
		\centering
		\includegraphics[width=0.45\textwidth]{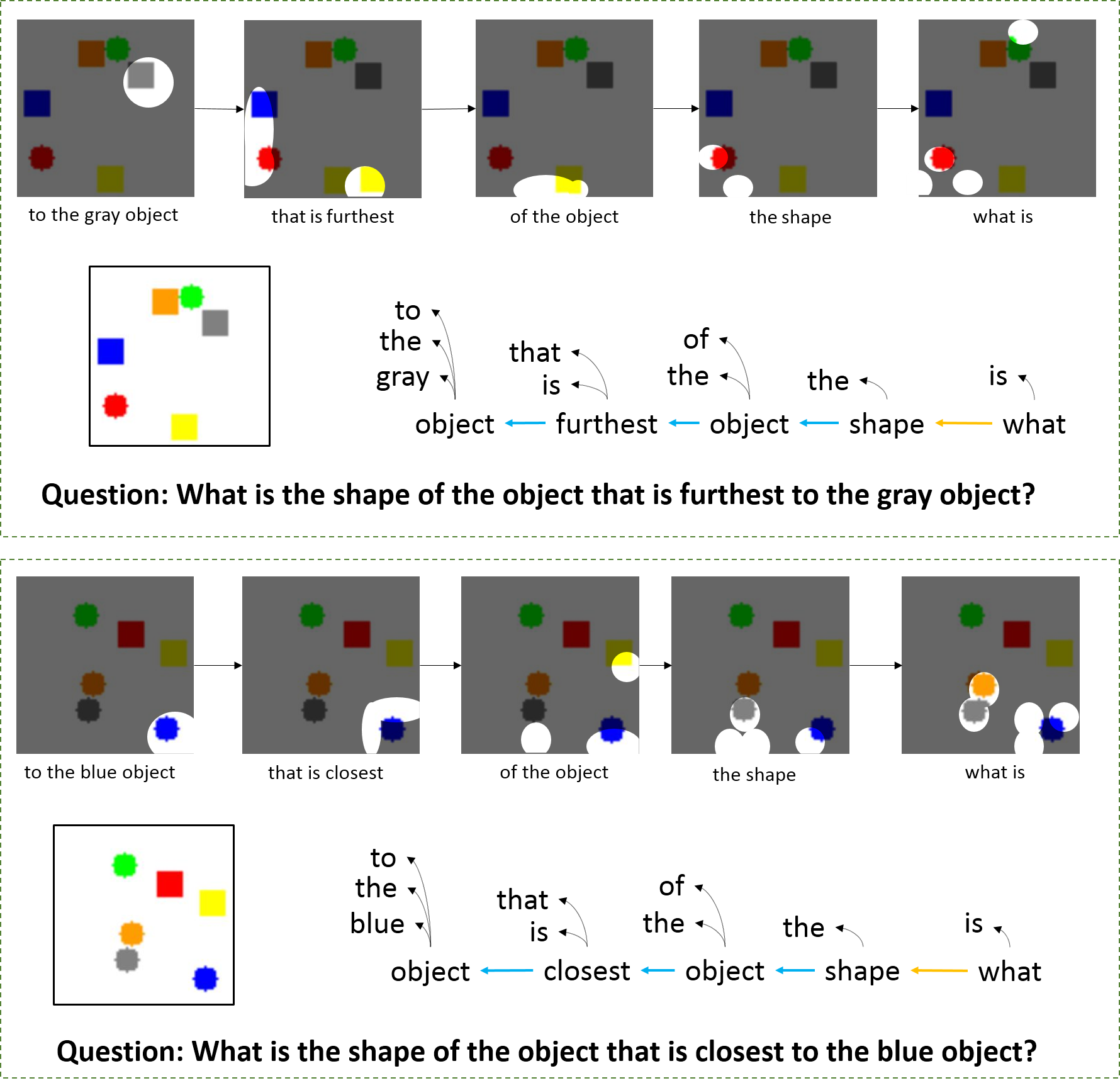}
		
		\caption{Examples of parse trees and corresponding regions attended by our ACMN on Sort-of-CLEVR dataset. Same with Figure~\ref{fig:clevr}, the edges in dependency tree is drawn from head words to dependent words. The attended regions are highlighted for different nodes. }
		\label{fig:sclevr}
		\vspace{-4mm}
	\end{figure}

	Figure~\ref{fig:sclevr} shows the resulting attention regions following the general dependency tree for the questions achieved by our ACMN, which clearly demonstrates its promising interpretable ability. The first example locates the ``gray object", then it transforms its attention regions to its ``furthest" objects. Our model successfully attends the correct objects in last steps to answer the question. The second example also attends the ``closet" area of ``blue object", and then correctly locate the gray circle object to answer the question.

	%\subsubsection{Visual7w dataset}
	%The results on Visual7W dataset are shown in Table~\ref{table:v7w}. The ``MLP (A, Q, I)~\cite{ECCV16baseline}" achieves the previous state-of-art performance, which takes the concatenated features of image, question, and a candidate answer from multiple choices as input, and classifies whether the answer is correct for the question. For fair comparison, we add soft-attention mechanism to their model by re-implementing this method with a few changes: 1) extract the image features from the conv5 of ResNet-152 instead of penultimate layer of ResNet-101 in their original model; 2) perform the soft attention on the image feature maps to obtain a $2048$-d features and concatenate it with the question and answer for prediction, named as ``MLP (A, Q, I) + att". It can be observed that incorporating the attention mechanism can help slightly improve the results of MLP (A, Q, I)~\cite{ECCV16baseline}. Our ACMN further outperforms ``MLP (A, Q, I)~\cite{ECCV16baseline}" and ``MLP (A, Q, I) + att" by $2.77\%$ and $1.84\%$ respectively, which demonstrates the rationality of integrating the adversarial attention module and residual composition module. 
	
	\subsubsection{VQAv2 dataset}
	The results on test-std and test-dev of the VQAv2 dataset are shown in Table~\ref{table:vqav2}. We compare our model with the first place method of the VQAv2 challenge. 
	The first place method obtained their best with an ensemble of 30 networks, and their results are denoted as ``1st ensemble~\cite{vqav2winner}" in Table~\ref{table:vqav2}. The ``1st single~\cite{vqav2winner}" show its performance of single network with exact same network architecture and hyper-parameters. Since they used the image features extracted by bottom-up attention network~\cite{Anderson2017up-down}, we also use features provided by \cite{Anderson2017up-down} for fair comparison. 
	Specifically, we use features of the top-$36$ proposal with highest object score as visual inputs and generate a $36$-D attention vector. 
	%then each proposal feature is adversarial masked by its child nodes. The masked features are fused with word vector, and generate its attention score separately. Finally, all $36$ scores are regularized with softmax to generate a $36$-D attention weight vector. 
	Our results are slightly lower than the best method on VQAv2. Note that we haven't applied tricks such as data augmentation, pretrained classifier, as described in \cite{vqav2winner}.

	\subsection{Ablation Studies} 
	We show the accuracy of our model and its variants on Sort-of-CLEVR dataset in Table~\ref{table:sclevr}. 
	
	\textbf{Residual composition module} By removing connection in our residual composition module, the accuracy drops $2.7\%$ on relational question answering by comparing ``Ours-w/o residual" with ``ours". Furthermore, we combine the residual and dense connection to form a dual-path tree-structured network, resulting in a variant ``Ours-DualPath". This network has more parameters and exploits previous nodes' knowledge in a more direct way. Specifically, we concatenate all previous hidden representation $h$ and use an extra fully connected layer to project them into a $256$-d feature vector. ``Ours-DualPath" achieves $91.1\%$ accuracy, indicates that the extra fully connected layer hurts the performance since nodes in a general dependency parse tree may contain duplicate information. Our residual connection can handle these trivial nodes, demonstrate the effectiveness of our residual composition. 
	
	%\begin{table}[!tp] \centering \small%\footnotesize %
	%	\center
	%	\resizebox{\columnwidth}{!}{%
	%		\begin{tabular}{|*{8}{c|}}
	%			\hline
	%			Method & What & Where & When & Who & Why & How & Overall \\
	%			\hline
	%			LSTM-Att~\cite{ECCV16baselineref} & 51.5 & 57.0 & 75.0 & 59.5 & 55.5 & 49.8 & 55.6 \\
	%			MCB + Att ~\cite{MCB} & 60.3 & 70.4 & 79.5 & 69.2 & 58.2 & 51.1 & 62.2 \\
	%			MLP (A, Q, I)~\cite{ECCV16baseline} & 64.5 & 75.9 & 82.1 & 72.9 & 68.0 & 56.4 & 67.1 \\
	%			MLP (A, Q, I) + att & 66.34 & \textbf{77.01} & \textbf{81.12} & \textbf{75.47} & 64.55 & 56.00 & 68.03 \\
	%			ours & \textbf{69.30} & 76.24 & 81.07 & 75.23 & \textbf{66.17} & \textbf{56.07} & \textbf{69.40} \\
	%			\hline
	%		\end{tabular}
	%	}
	%	\caption{The question answering accuracy on Visual7W dataset. Each column shows the accuracy of a particular question type.}
	%	\label{table:v7w}
	%	\vspace{-3mm}
	%\end{table}

	\begin{table}[!tp] \centering %\small%\footnotesize %
		\center
		\resizebox{\columnwidth}{!}{%
			\begin{tabular}{|c|*{4}{c}|*{4}{c}|}
				\hline
				& \multicolumn{4}{c|}{test-dev} & \multicolumn{4}{c|}{test-std} \\
				Method & All & Yes/no & Numb. & Other & All & Yes/no & Numb. & Other\\
				\hline
				1st ensemble~\cite{vqav2winner} & 69.87 & 86.08 & 48.99 & 60.80 & 70.34 & 86.60 & 48.64 & 61.15 \\
				%1st ResNet features 7×7~\cite{vqav2winner} & 62.07 & 79.20 & 39.46 & 52.62 & 62.27 & 79.32 & 39.77 & 52.59 \\
				1st single~\cite{vqav2winner} & 65.32 & 81.82 & 44.21 & 56.05 & 65.67 & 82.20 & 43.90 & 56.26 \\
				ours & 63.81 & 81.59 & 44.18 & 53.07 & 64.05 & 81.83 & 43.80 & 53.22 \\
				\hline
			\end{tabular}
		}
		\caption{The question answering accuracy on VQAv2 test-dev and test-std. The ``1st ensemble" and ``1st single" denotes the first place method of the 2017 VQA Challenge with and without ensemble respectively.}
		\label{table:vqav2}
		\vspace{-4mm}
	\end{table}
	
	\textbf{Adversarial attention module} We also evaluate the results of other attention modules to demonstrate the effectiveness of our adversarial attention module. One commonly used attention module is the Relocate module in \cite{e2emn} which used the soft-attention encoding applied in \cite{e2emn}, resulting in our variant ``Ours-relocate". Another option for attention module is to directly concatenate image features with the input attention maps $att_{in}$ instead of using an adversarial mask, that is ``Ours-concat". The proposed adversarial attention module is demonstrated to obtain better question answering performance over these two attention alternatives, benefiting from the adversarial-mask driven exploration of unseen regions.
	
	\section{Conclusion}
	In this paper, we propose a novel ACMN module network equipped with an adversarial attention module and a residual composition module for visual question reasoning. In contrast to previous works that rely on the annotations or hand-crafted rules to obtain valid layouts, our ACMN model can automatically perform interpretable reasoning process over a general dependency parse tree from the question, which can largely broaden its application fields. The adversarial attention module encourages the model to attend the local visual evidence for each modifier relation while the residual composition module can learn to compose representations of children for the clausal predicate relation while retaining the information flow from its indirect child nodes. Experiments show that our model outperforms previous modular networks without using any specified groundtruth layouts or complicated hand-crafted rules.
	%Our model further achieves superior results on natural image datasets and generates the interpretable intermediate reasoning results.
	\clearpage
	
	{\small
		\bibliographystyle{ieee}
		\bibliography{egbib}
	}
	
\end{document}